\documentclass{article}

\usepackage[final]{neurips_2021}
\usepackage{hyperref}
\hypersetup{hidelinks,
breaklinks=true,
colorlinks=true,
urlcolor=blue,
bookmarksopen=false,
pdftitle={Title},
pdfauthor={Author}}

\usepackage[utf8]{inputenc} 
\usepackage[T1]{fontenc}    
\usepackage{hyperref}       
\usepackage{url}            
\usepackage{booktabs}       
\usepackage{amsfonts}       
\usepackage{nicefrac}       
\usepackage{microtype}      
\usepackage{xcolor}         
\usepackage{mathtools}
\usepackage{amssymb}
\usepackage{amsmath}
\usepackage{caption}
\usepackage{subcaption}
\usepackage{multirow}

\newcommand{\comment}[1]{}
\DeclarePairedDelimiterX{\inner}[2]{\langle}{\rangle}{#1, #2}

\usepackage{changepage}
\newlength{\offsetpage}
{\end{adjustwidth}}

\title{Using Early-Learning Regularization to Classify Real-World Noisy Data}

\author{
  Alessio Galatolo\textsuperscript{$*$}
  \quad
  Alfred Nilsson\textsuperscript{$\dagger$}
  \quad
  Roderick Karlemstrand\textsuperscript{$\ddagger$}
  \quad
  Yineng Wang\textsuperscript{$\mathsection$} \footnotemark[5]

  \\\\
  $*$ M.Sc. student in Machine Learning \quad
  $\dagger$ M.Eng. student in Engineering Physics \quad \\
  $\ddagger$ M.Eng. student in Information Technology \quad
  $\mathsection$ M.Sc. student in Computer Science \quad \\
  \\
  School of Electrical Engineering and Computer Science, \\
  KTH Royal Institute of Technology
  \\
  \small{\texttt{\{galatolo, alfredn, karlemst, yineng\}@kth.se}}
}

\usepackage[style=ieee]{biblatex}
\begin{document}

\maketitle

\begin{abstract}

The memorization problem is well-known in the field of computer vision. Liu et al. propose a technique called Early-Learning Regularization, which improves accuracy on the CIFAR datasets when label noise is present. This project replicates their experiments and investigates the performance on a real-world dataset with intrinsic noise. Results show that their experimental results are consistent. We also explore Sharpness-Aware Minimization in addition to SGD and observed a further 14.6 percentage points improvement. Future work includes using all 6 million images and manually clean a fraction of the images to fine-tune a transfer learning model. Last but not the least, having access to clean data for testing would also improve the measurement of accuracy.

\end{abstract}

\renewcommand{\thefootnote}{\fnsymbol{footnote}}
\footnotetext[5]{All authors contributed equally to this work.}
\renewcommand*{\thefootnote}{\arabic{footnote}}

\section{Introduction}

Supervised image classification is usually done on clean and well-structured datasets. If the labels include noise, neural networks tend to overfit the noisy labels. This is known as the memorization effect. However, Liu et al. demonstrate in their 2020 NeurIPS paper that it is possible to prevent memorization on the well-known CIFAR-10 and CIFAR-100 datasets by introducing a new regularization technique \cite{liu2020early}.

In this project, we validate the results of Liu et al. on the aforementioned datasets and then extend this regularization technique to a real-world dataset provided by CDON.

CDON is a Swedish e-commerce company that operates throughout Scandinavia. Their catalog is vast and varied as they count millions of products among hundreds of categories. CDON also functions as a marketplace and is open to third-party sellers from which the majority of the products originate. However, since the insertion of these products is left entirely to the seller, products are not always placed in the correct categories, which may confuse the end-user. Their development team estimates that the percentage of incorrectly classified products is between 10\% and 20\%. They consider this as a primary issue to solve as they are currently investing in manual corrections, which is slow and expensive. Our goal is to use the methods proposed in Liu et al.'s paper to train a deep neural network that can correctly classify the products.

\section{Related Work}

Image classification with noisy labels is a well-known subject in Machine Learning research and there are many different approaches in understanding the learning of noisy labels. Xingjun et al. investigated the use of two different loss functions that together avoid the effect of underfitting while also being robust towards noisy labels \cite{ma2020normalized}. Wang et al. explored using classic cross-entropy loss symmetrically with a noise-robust counterpart, Reverse Cross-Entropy, to both decreased the overfitting and the under-learning \cite{wang2019symmetric}. Liu et al. showed how a neural network trained with noisy labels tended to memorize such labels but only on advanced stages of the training process \cite{liu2020early}. For this reason, they designed a regularization term to try and avoid the memorization issue in the later stages of training.
Among these, the Early-Training Regularization (ELR) method is the one that showed the best results in real-world datasets~\cite{papers2020code} and, therefore, is the one we implement in our work. Furthermore, many of the best result on CIFAR-100 classification used SAM (Sharpness-Aware Minimization). SAM computes the gradients in the first forward pass and tries to smooth them to improve final prediction accuracy \cite{foret2020}.

\quad

\section{Data}

For this project, we use three datasets. The first two are CIFAR-10 and CIFAR-100, which consist of 60 000 images of 10 and 100 classes respectively.

The last dataset is generated by CDON's database. CDON provided, for each product category, a file with all the products in that category alongside their details\footnote{The categories at CDON are organized as a tree, with 21 main categories such as `home electronics' or `clothing' and many subcategories. `CPU (processor)' and `mice' are examples of subcategories for `home electronics'. Further information can be found at \hyperlink{https://cdon.se/}{cdon.se}.}. We use the thumbnail of the product as the image for our classification and the subcategory ID as the label. We filtered out the invalid products and those with an old release date.

The number of different subcategories and the size of the dataset greatly exceed our expectation as we are able to count more than 4 000 different subcategories and over 6 million products. After an in-depth analysis of the distribution of the subcategories, we discover that most of them (over 90\%) have only a few samples, with the other ones holding most of the products. For this reason, we choose to only use those large categories. Due to our limited time frame and computing power, we also decide on using only a portion of it. We use 63 000 images evenly distributed among 63 subcategories. Each subcategory has 1 000 samples. They are then divided into a training set and a test set with a ratio of 9:1. We do not have any completely clean dataset available. Although this is expected, most of the state-of-the-art methods were evaluated on real-world data, still have a limited, though meaningful, clean dataset to test on.

\quad

\section{Methods}

Because our goal was to replicate the results of Liu et al. we adapted the same network architecture, ResNet-34, and their novel regularization technique ELR.

\subsection{Residual Neural Network: ResNet}

A residual neural network, ResNet for short, utilizes \textit{blocks} and biologically inspired "skip connections" or identity mappings \cite{kaiming2015resnet}. Each block consist of a series of convolutional layers and each block has a skip connection from the input to the block to the output of the block, see figure \ref{fig:resnet-18}.

This design is an attempt to deal with the \textit{degradation} problem of deep neural networks. While the issue of vanishing gradients is mostly solved through the use of batch normalization, degradation refers to the saturation of the accuracy the deeper the networks get. Degradation is not the result of overfitting, but rather a reflection of the fact that all systems are not equally easy to optimize. Complex models with multiple non-linear layers might have difficulties in approximating simple identity mappings, and in situations where the identity mapping is optimal, gradient-descent can force the weights of the non-linear layers toward zero to approximate the identity mapping.

The ResNet architecture has been shown empirically to provide stable solutions and excellent results for hundreds of layers \cite{kaiming2015resnet}.

\subsection{ResNet-34}

\begin{figure}[ht]
  \centering
  \includegraphics[width=1.0\linewidth]{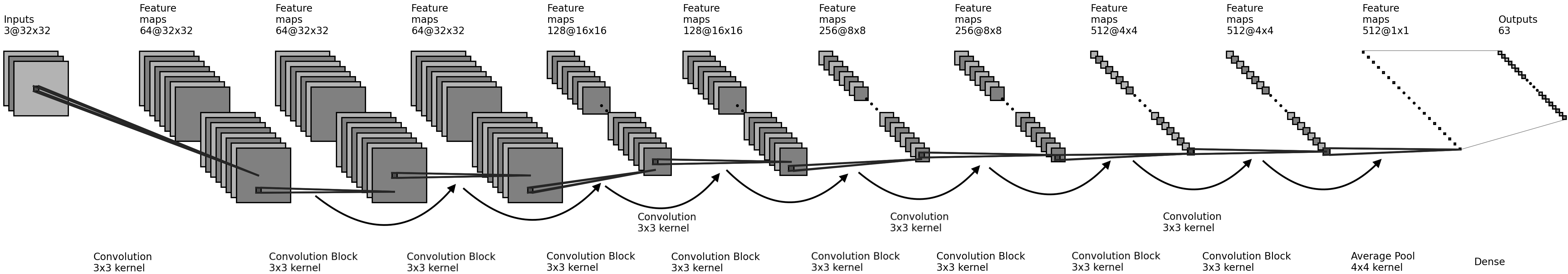}
  \caption{A ResNet-18 network architecture with 63 output classes}
  \label{fig:resnet-18}
\end{figure}

An 18-layer ResNet architecture is illustrated in figure \ref{fig:resnet-18}. The network consists of six parts: an input layer, four residual blocks and an output layer. Inside each residual block, there are two convolutional blocks.
Our experiments focused mostly on the 34-layer version called ResNet-34 which tries to identify more complex patterns in the data and also consists of four residual blocks, but which contain 3, 4, 6 and 3 convolutional blocks respectively.

\subsection{Early-Learning Phenomenon and Regularization}
\label{ssec:elr-phenomenon}
\paragraph{Memorization}
When training on data where noise is present in the labels, deep neural networks have been observed to overfit the noisy labels.
This is defined by the authors as "memorization". During the early phase of learning, they tend to learn the true labels and move toward the true optimum, before they start memorizing the incorrect labels and start approaching a different false optimum.

While this proof assumes a linear classifier, the authors go on to show that the effect of label noise on deep neural classifiers is analogous.
Thus, the theorem suggests that we need to find a way to keep the gradients corresponding to correctly labelled samples large while preventing the gradients corresponding to samples with incorrect labels from dominating as the training goes on.

\paragraph{Early-Learning Regularization}
Sheng Liu et al. propose a new regularization designed to combat memorization of noisy labels, which they call "Early-Learning Regularization". The regularization is added as an extra term to the Cross-Entropy loss function, see the mathematical description in equation \ref{eq:ELR}.
\begin{equation}
    \mathcal{L}_\text{ELR}(\Theta) := \mathcal{L}_\text{CE}(\Theta) + \frac{\lambda}{n} \sum_{i=1}^{n} \log \left(1 - \inner{\mathbf{p}^{(i)}}{\mathbf{t}^{(i)}} \right)
    \label{eq:ELR}
\end{equation}
where $\Theta$ is the network parameters, $\mathbf{p}^{(i)}$ is the softmax probabilities corresponding to sample $i$ and $\mathbf{t}^{(i)}$ is a moving average of all previous activations of the last layer corresponding to a sample $i$.

This regularization term exploits the early learning phenomenon. It places a penalty on outputs $\mathbf{p}$ that deviate from the moving average $\mathbf{t}$ of previous activations by maximizing the inner product $\inner{\mathbf{p}^{(i)}}{\mathbf{t}^{(i)}}$. It is assumed that the early iterations of training move the weights toward the correct minimum, we penalize activations that are too far (distance measured by the inner product) from the average of the initial activations, to counter-act memorization of noisy labels.

Further understanding can be obtained by examining the gradient of $\mathcal{L}_\mathrm{ELR}(\Theta)$, but we refer to the original paper for further detail \cite{liu2020early}.

\quad

\section{Experiments}

In this section, we investigate Early-Learning Regularization on benchmark datasets CIFAR-10 and CIFAR-100, and the dataset provided by CDON.

\subsection{The Early-Learning Phenomenon on the CIFAR-10 Dataset}

\begin{figure}[htb!]
  \centering
  \subfloat[Cross entropy loss]{\includegraphics[scale=0.36]{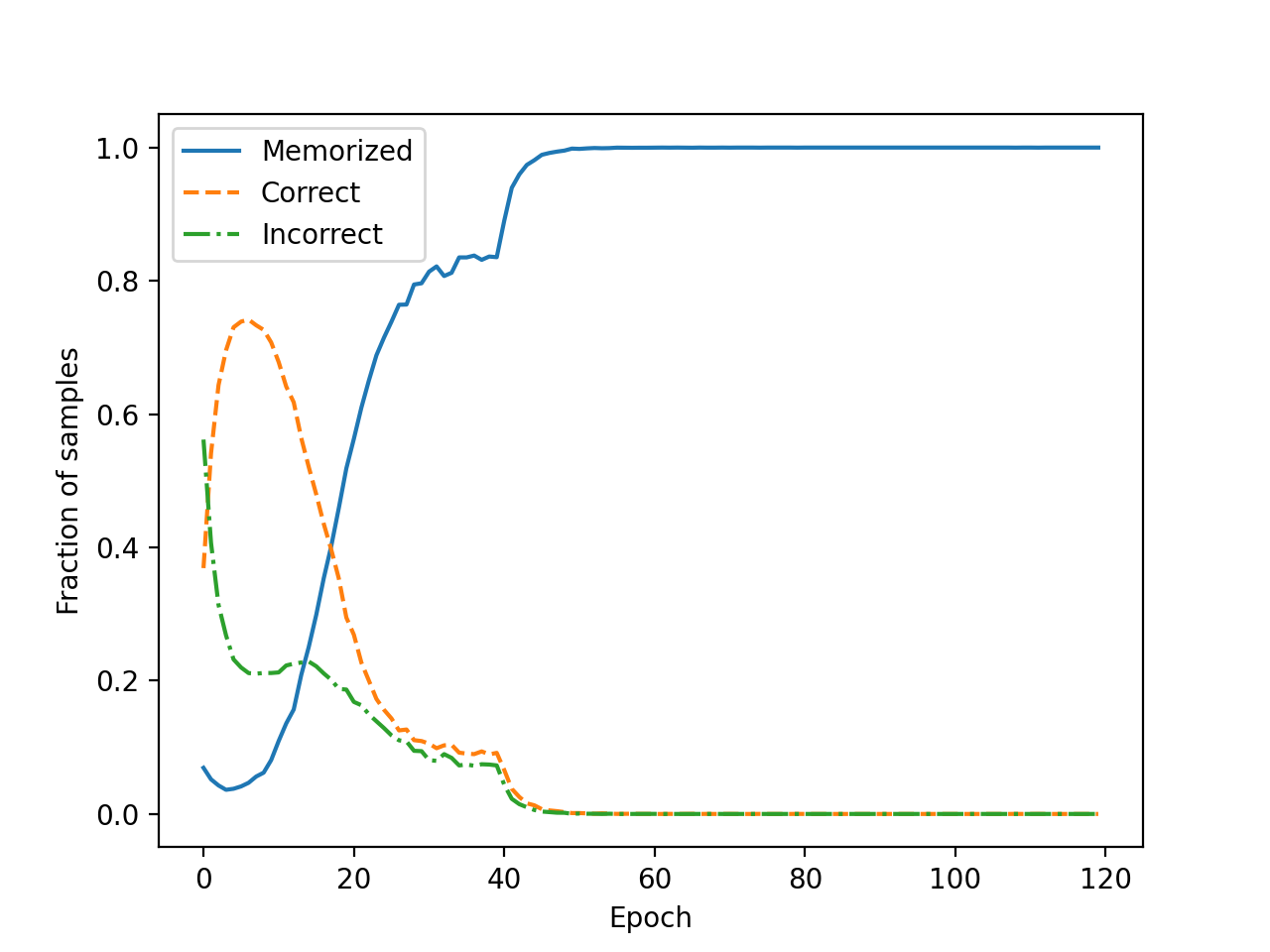}}
  \subfloat[ELR]{\includegraphics[scale=0.36]{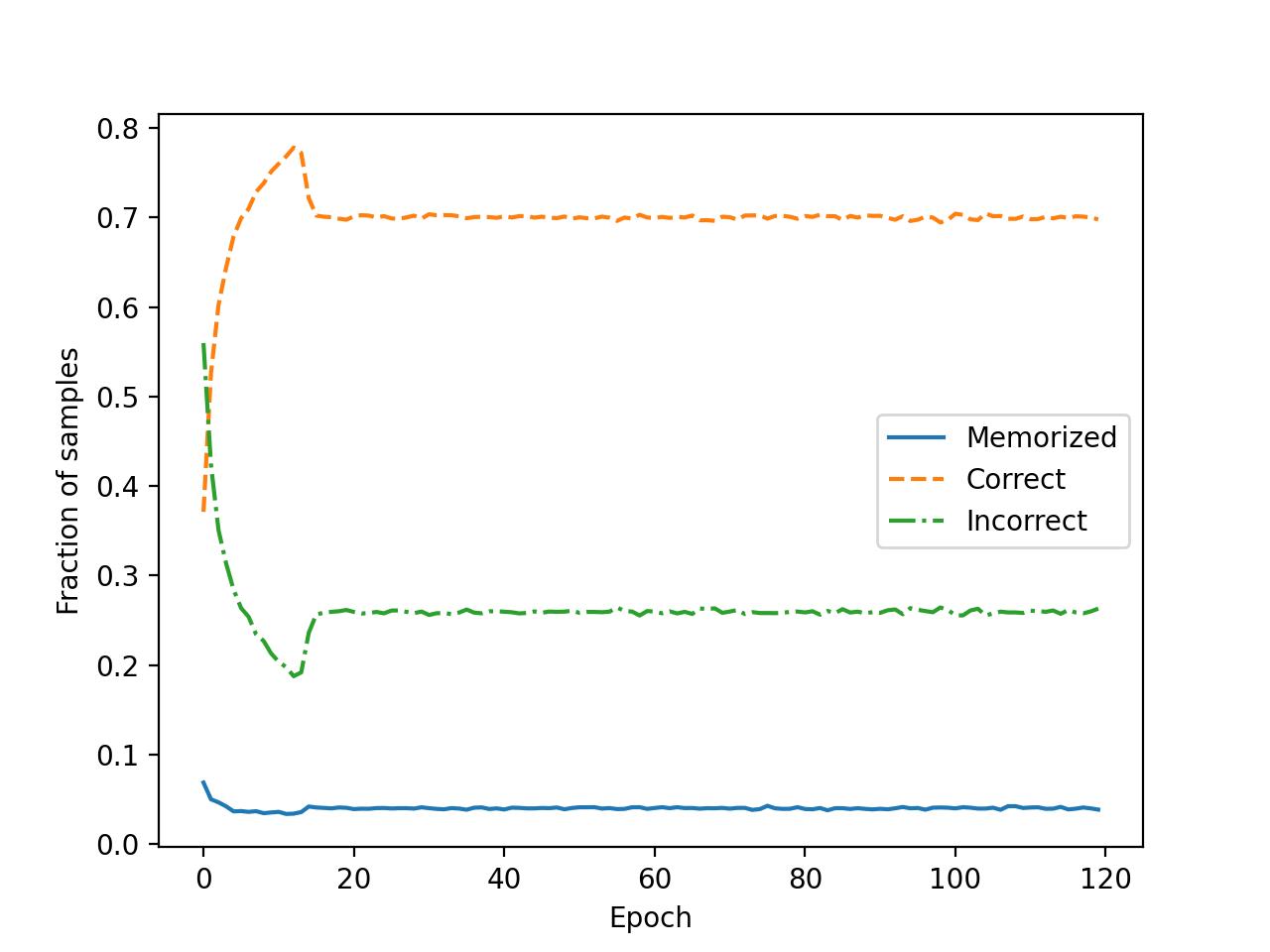}}
  \caption{Illustration of the early-learning phenomenon}
  \label{fig:mem_ce}
\end{figure}

To confirm the early-learning phenomenon proposed by the original paper and discussed in Section~\ref{ssec:elr-phenomenon}, we train on CIFAR-10 with 20\% manually added symmetric noise, and record the fraction of training samples with wrong labels which are predicted correctly (as the true label), memorized (predicted as the labeled class), and incorrectly predicted as neither the true label nor the labeled class. We observe the early-learning phenomenon using only cross-entropy loss, as is presented in Figure~\ref{fig:mem_ce} (a). The model is observed to fit the clean data in the early training stage but memorized almost all samples with wrong labels by epoch 50. By using early-learning regularization, memorization is observed within only a small fraction of the samples.

\subsection{Evaluation on CIFAR-10/100 datasets}

\paragraph{Data pre-processing}
We apply normalization, random crop and horizontal flip on the training sets. For CIFAR-10 and CIFAR-100, we manually add noise to the labels by using the scheme of generating symmetric noise proposed in the original paper. The noisy labels are simulated by flipping a proportion of the labels following a symmetric uniform distribution. We choose three different noise levels, 10\%, 15\%, and 20\%, along with the case no noise is added.

\paragraph{Networks and training}
We train the networks using Cross-Entropy (CE) Loss as the baseline to evaluate the performance of ELR. To replicate the results achieved by the original paper, the same model is used. For all training, we use an SGD optimizer with momentum 0.9, weight decay 0.001, and a batch size of 128. We train the networks for 120 epochs for CIFAR-10, and 150 epoch for CIFAR-100. Two different schemes of learning rate annealing are used. For CIFAR-10, we choose the initial learning rate as 0.02 and decrease it by 100 after 40 and 80 epochs, and after 80 and 120 epochs for CIFAR-100. Another scheme is cosine annealing learning rate, for which we set the maximum number of epochs for each period to 10, and the minimum and maximum learning rate to 0.001 and 0.02, respectively. We also choose the same temporal ensembling parameter $\beta$ and regularization coefficient $\lambda$ as the original paper, which are $\beta=0.7$ and $\lambda=3$ on CIFAR-10, as well as $\beta=0.9$ and $\lambda=7$ on CIFAR-100.

\paragraph{Results}
The classification accuracies using CE loss and ELR are reported in Table~\ref{tbl:cifar-basic-results}. As is shown, our implementation of ELR outperforms CE loss by a significant margin for noise level 10\%, 15\% and 20\%. We also compare the performance when the noise level is 0, and ELR does not produce an improvement in accuracy on CIFAR-100, which indicates ELR is likely to not have a considerable impact on the test accuracy when all labels are clean.

\begin{table}
  \caption{Test accuracies (\%) using CE loss and ELR with symmetric noise on CIFAR-10/100}
  \label{tbl:cifar-basic-results}
  \centering
  \begin{tabular}{c|c|cc|cc}
    \toprule
    \multirow{2}{*}{\begin{tabular}[c]{@{}c@{}}Datasets\\ (Architecture)\end{tabular}} & \multirow{2}{*}{\begin{tabular}[c]{@{}c@{}}Symmetric\\
    label noise\end{tabular}} & \multicolumn{2}{c|}{\small{Multi-step LR}} & \multicolumn{2}{c}{\small{Cosine annealing}} \\
    & & \small{CE} & \small{ELR} & \small{CE} & \small{ELR} \\
    \midrule

    \multirow{4}{*}{\begin{tabular}[c]{@{}c@{}}CIFAR-10\\(ResNet-34)\end{tabular}}
    & 0 & 83.84 & 88.94 & 82.03 & 89.32 \\
    & 10\% & 78.55 & 87.04 & 75.05 & 87.08 \\
    & 15\% & 78.88 & 86.35 & 77.85 & 85.72 \\
    & 20\% & 72.30 & 84.85 & 74.07 & 83.50 \\
    \midrule
    \multirow{4}{*}{
    \begin{tabular}[c]{@{}c@{}}CIFAR-100\\(ResNet-34)\end{tabular}}
    & 0 & 75.34 & 73.66 & 75.48 & 74.79 \\
    & 10\% & 68.52 & 73.09 & 68.79 & 73.13 \\
    & 15\% & 65.34 & 71.94 & 64.99 & 72.86 \\
    & 20\%  & 61.82 & 71.10 & 61.41 & 71.9 \\
    \bottomrule
  \end{tabular}
\end{table}

\subsection{Evaluation on CDON's dataset}

\paragraph{Data pre-processing}
The images of products provided by CDON are of different sizes. Most of them have $80 \times 80$ pixels, but some vary by several pixels in width and/or height. We resize the images to $32 \times 32$ pixels and then pre-process the data similarly.

\paragraph{Networks and training}
We use the ResNet-34 network architecture. We randomly split the original dataset into training and test sets, with a ratio of 9:1. The selected values for ELR hyper-parameters are $\beta=0.9$ and $\lambda=7$. Cosine annealing learning rate is used. The other parameter settings are the same as the experiments for CIFAR-100.

\begin{figure}[htb!]
  \centering
  \includegraphics[width=\linewidth]{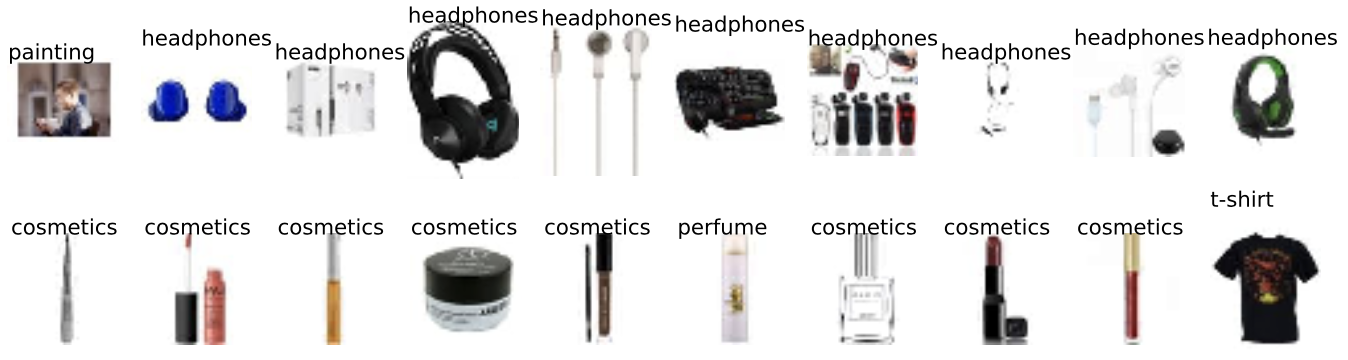}
  \caption{
  Examples of products in CDON's dataset, where the bottom-right image of a T-shirt is the only noise data. Products in the first row are all labeled as ``earphone'', and the second row labeled as ``cosmetic''. Above the images are our model's predictions. The top-left image predicted as ``painting'' depicts a person wearing earphones, which causes interference to the prediction. The bottom-right image predicted as ``t-shirt'' presents a real T-shirt and is an example of noisy data.}
  \label{fig:cdon-prediction}
\end{figure}

\paragraph{Results}
We conduct a preliminary examination of the prediction results on CDON's dataset using ELR. Figure~\ref{fig:cdon-prediction} presents several examples. Since noisy labels are inherent in CDON's dataset, it is impossible to precisely compute the accuracy concerning clean data. We still compute the fraction of samples that are predicted as their original categories as the test accuracy in further experiments, though this introduces slight biases to the true accuracy.

\subsection{Hyper-parameters Fine-tuning}

\begin{figure}[ht]
  \centering
  \includegraphics[width=1.0\textwidth]{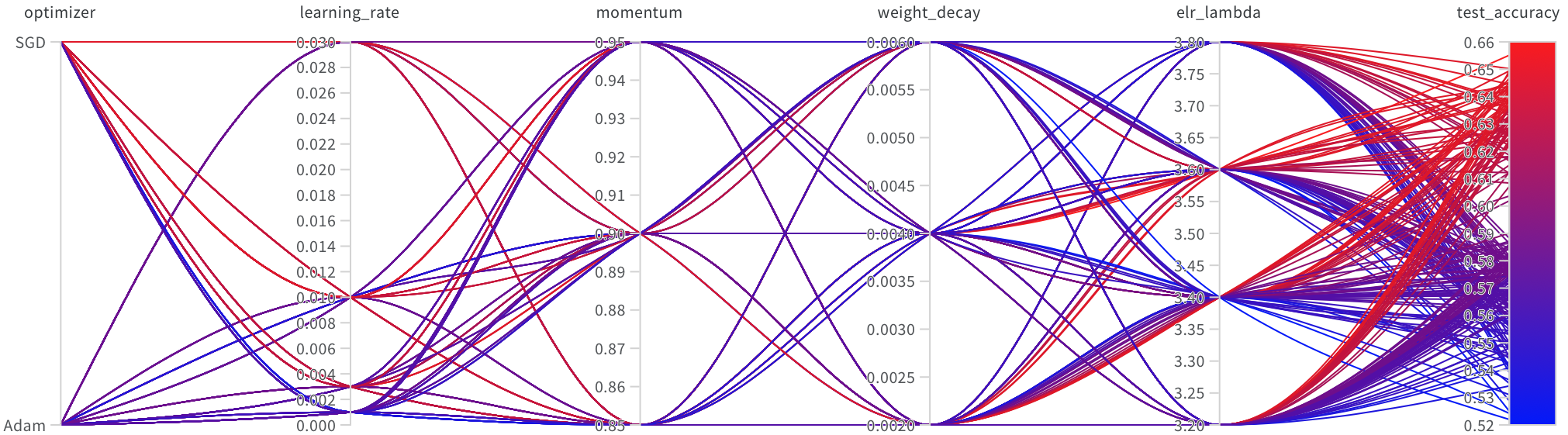}
  \caption{The second sweep, a grid search of 5 hyperparameters, where each run tries different combinations of the hyperparameters and the model is trained for 30 epochs. The best combination is used for further training for 150 epochs on CDON dataset.}
  \label{fig:sweep}
\end{figure}

We fine-tune our model with a random search over 8 hyperparameters and a grid search using a technique called a sweep. The two sweeps try out all over 400 possible combinations over 8 dedicated GPUs with a total computing time of over 300 hours. Figure \ref{fig:sweep} shows the mapping between combinations of hyperparameters and test accuracy.

\begin{figure}[ht]
    \centering
    \begin{subfigure}[b]{0.45\textwidth}
        \includegraphics[width=\textwidth]{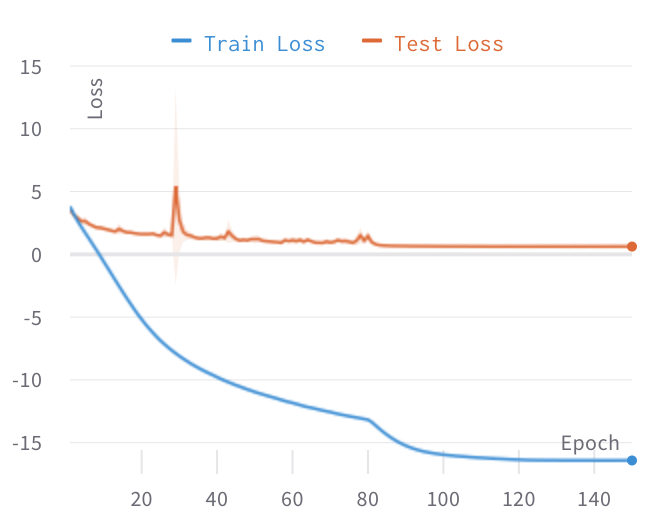}
        \caption{The training and test loss}
        \label{fig:exp19-a}
    \end{subfigure}
    \qquad
    ~ 
    \begin{subfigure}[b]{0.45\textwidth}
        \includegraphics[width=\textwidth]{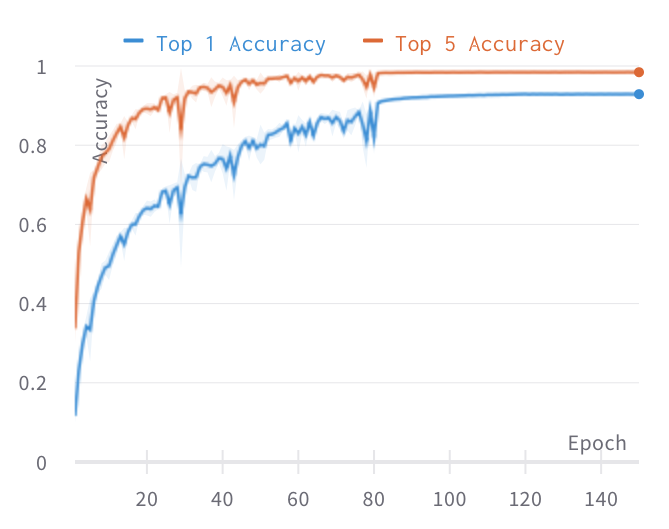}
        \caption{The top 1 and top 5 accuracy on the test set}
        \label{fig:exp19-b}
    \end{subfigure}
  \caption{The ResNet-34 model with CE and ELR, trained with SGD and SAM using fine-tuned hyperparameters and Multi-step Learning Rate scheduler. The training loss is measured with CE and test loss is measured with ELR. Results are from 4 independent runs with an estimated total computation of 1 958 400 TFLOPS. The figures show mean and standard deviations.}
  \label{fig:exp19}
\end{figure}

We find among others that $\lambda$ value 3.6 and $\beta$ 0.85 give the best results after 30 epochs. We then train our model for 150 epochs. The final test accuracy reaches 78.8\%. We also incorporate SAM into our SGD learning algorithms. The final top 1 result has a mean value of 93.4\% and a variance of 0.003. The top 5 results reach 98.2\%. Figure \ref{fig:exp19} shows the results of 4 runs. In each run, the dataset is shuffled and training data and test data is divided afterwards.

\quad

\section{Conclusion}

By utilizing an Early-Learning Regularization term we are able to considerably improve the performance of our network in the case of noisy data. We have verified that using this method in CIFAR-10/100 improves the accuracy compared to a plain CE. We also see an improvement on the CDON dataset although limited. Future work could probably include more in-depth training using far more samples than we did and possibly getting an accuracy high enough to be actively used by CDON. One possible application is to put this Neural Network as a preliminary check when a new product is inserted by a third-party seller. If the network's classification differs from the proposed one, a human can double-check and assess the right category. Although this is not a conclusive solution, it could still save a lot of time in the processing of new products.

\begin{ack}
This project is sponsored by CDON. We would also like to thank Linode for supporting us with three Nvidia Quadro RTX 6000 and Google for supporting us with a Tesla V100 and a Tesla T4.
\end{ack}

\newpage

\printbibliography

\end{document}